\documentclass[10pt,twocolumn,letterpaper]{article}

\usepackage{wacv}
\usepackage{times}
\usepackage{epsfig}
\usepackage{graphicx}
\usepackage{amsmath}
\usepackage{amssymb}
\usepackage[accsupp]{axessibility} 

\usepackage{url}            
\usepackage{booktabs}       
\usepackage{amsfonts}       
\usepackage{nicefrac}       
\usepackage{microtype}      
\usepackage{authblk}
\usepackage{graphicx}
\usepackage{amssymb}
\usepackage{amsmath}
\usepackage{color}
\usepackage{threeparttable}
\usepackage{lipsum}
\usepackage{enumitem}

\def\wacvPaperID{11} 

\wacvfinalcopy 

\ifwacvfinal
\fi

\ifwacvfinal
\usepackage[breaklinks=true,bookmarks=false]{hyperref}
\else
\usepackage[pagebackref=true,breaklinks=true,colorlinks,bookmarks=false]{hyperref}
\fi

\begin{document}
\graphicspath{{images/}}
\title{Argus++: Robust Real-time Activity Detection\\for Unconstrained Video Streams with Overlapping Cube Proposals}

\author{Lijun Yu, Yijun Qian, Wenhe Liu, and Alexander G. Hauptmann\\
Language Technologies Institute, Carnegie Mellon University\\
{\tt\small \{lijun, yijunqian@cmu.edu\}, \{wenhel, alex\}@cs.cmu.edu}
}

\maketitle
\thispagestyle{empty}

\begin{abstract}
    Activity detection is one of the attractive computer vision tasks to exploit the video streams captured by widely installed cameras.
Although achieving impressive performance, conventional activity detection algorithms are usually designed under certain constraints, such as using trimmed and/or object-centered video clips as inputs.
Therefore, they failed to deal with the multi-scale multi-instance cases in real-world unconstrained video streams, which are untrimmed and have large field-of-views.
Real-time requirements for streaming analysis also mark brute force expansion of them unfeasible.

To overcome these issues, we propose \textit{Argus++}, a robust real-time activity detection system for analyzing unconstrained video streams. 
The design of \textit{Argus++} introduces \textit{overlapping spatio-temporal cubes} as an intermediate concept of activity proposals to ensure coverage and completeness of activity detection through over-sampling.
The overall system is optimized for real-time processing on standalone consumer-level hardware.
Extensive experiments on different surveillance and driving scenarios demonstrated its superior performance in a series of activity detection benchmarks, including CVPR ActivityNet ActEV 2021, NIST ActEV SDL UF/KF, TRECVID ActEV 2020/2021, and ICCV ROAD 2021.


\end{abstract}

\section{Introduction}

\begin{figure*}[!t]
	\centering
	\includegraphics[width=\linewidth]{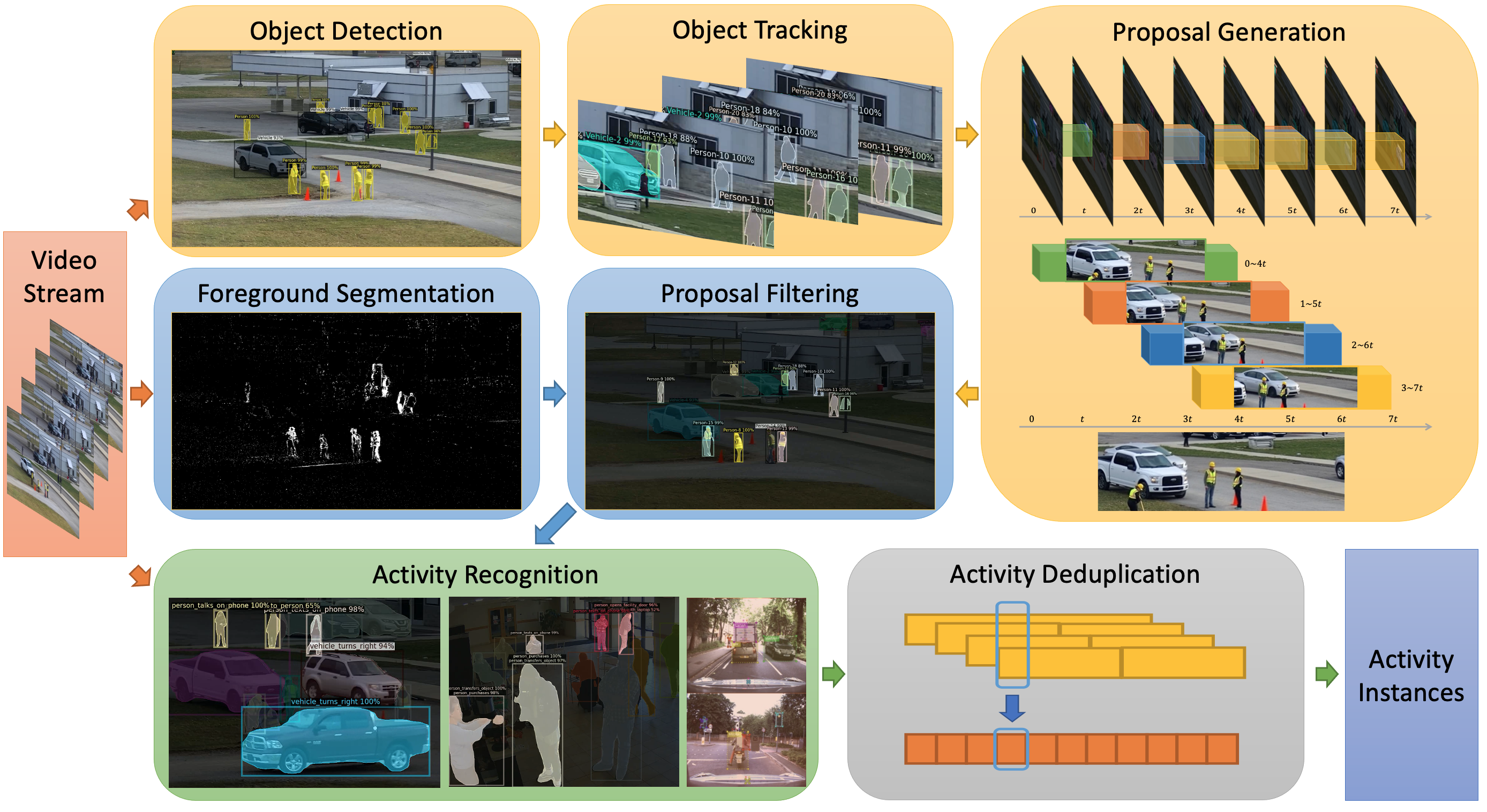}
	\caption{Architecture of \textit{Argus++}. A video stream is processed frame-by-frame through object detection and tracking to generate overlapping cube proposals. With frame-level foreground segmentation, stable proposals are filtered out. Activity recognition models determine the classification scores for each proposal. These over-sampled cubes are deduplicated to produce the final activity instances. }
	\label{fig/arch}
\end{figure*}

Nowadays, activity detection has drawn a fast-growing attention in both industry and research fields. Activity detection in extended videos \cite{corona_meva_2021,oh_large-scale_2011} is widely applied for public safety in indoor and outdoor scenarios. Activity detection on streaming videos captured by in-vehicle cameras is applied for vision-based autonomous driving. The development of these applications brings several challenges. First, most of these systems take \textit{unconstrained} videos as input, which are recorded in large field-of-views where multi-object and multi-activity occur simultaneously and continuously over time. Second, the unconstrained videos in real world are in multiple scenarios and under multiple conditions, e.g. in dynamically changed road environments from day to night in autonomous driving \cite{singh_road_2021}.  Third, efficient algorithms are 
demanded for real-time processing and responding of streaming video.

Conventional activity detection works \cite{tran_closer_2018,feichtenhofer_x3d_2020,wang_temporal_2016,kay_kinetics_2017,ghadiyaram_large-scale_2019} have achieved impressive performance. However, they are not suitable for real world unconstrained video understanding. Most of these works are applied under certain constrains, e.g., only for processing trimmed and/or object-centered video clips. Meanwhile, they usually are specified for certain scenarios, such as person activity, etc. Therefore, such algorithms would fail when being transferred to unconstrained videos on both efficiency and effectiveness.

Previous works \cite{rizve_gabriella_2021,yu_cmu_2020,liu_argus_2020} on unconstrained video analysis proposed to generate and analyze tube/tubelet proposals, which are trajectories extracted from object detection and tracking results. Tube proposal has several drawbacks. 
First, tube proposals failed to capture the trace of moving objects when cropping the proposals from the original videos. Therefore, learning the activities highly relied on trace would be difficult, e.g. 'vehicle turning right'.
Second, the tube proposals still cannot stay away from temporal activity localization to determine the existence of the activities.
Besides, most of the previous works \cite{rizve_gabriella_2021} utilize non-overlapping proposals, which straightforwardly cuts the tube proposals by fixed length of temporal windows. Inevitably, such methods destroy the completeness of activities. Therefore, it would result in significant degrade of performance.
Third, the objects in the tube proposal will suffer from the bounding box shift and distortion across frames, which could result in a high false alarm rate on activity detection.

To overcome the aforementioned challenges,  we propose \textit{Argus++}, an efficient robust spatio-temporal activity detection system for extended and road video activity detection. The proposed system contains four-stages: Proposal Generation, Proposal Filtering, Activity Recognition and Activity Deduplication.  The major difference between \textit{Argus++} and the former works, such as \cite{liu_argus_2020}, is the concept of \textit{cube} proposals. Rather than simply adapted tube proposals, i.e. cropped trajectories of detected and tracked objects, we propose to merge and crop the area of detected objects across the frames.

We summarize the contributions of our work as follows:
\begin{enumerate}[nosep]

    \item We propose \textit{Argus++}, a real-time activity detection system for unconstrained video streams, which is robust across different scenarios.
    \item We introduce overlapping spatio-temporal cubes as the core concept of activity proposals to ensure coverage and completeness of activity detection through over-sampling.
    \item The proposed system has achieved outstanding performance in a large series of activity detection benchmarks, including CVPR ActivityNet ActEV 2021, NIST ActEV SDL UF/KF, TRECVID ActEV 2020/2021, and ICCV ROAD 2021.
\end{enumerate}

\section{Related Work}

\paragraph{Object Detection and Tracking}
Object detection and tracking are fundamental computer vision tasks that aims to detect and track objects from images or videos. 
Image-based object detection algorithms, such as Faster R-CNN~\cite{ren_faster_2015} and R-FCN~\cite{dai2016r}, have demonstrated convincing performance but are often expensive to apply on every frame. 
Video-based object detection algorithms~\cite{zhu_flow-guided_2017, qian_adaptive_2020} use optical flow guided feature aggregation to leverage motion information and reduce computation.
With the deep features extracted from the backbone convolutional network, multi-object tracking algorithms~\cite{wojke2017simple, vedaldi_towards_2020} associates objects across frames based on feature similarity and location proximity.

\paragraph{Activity Detection}
In recent years, there emerged some systems designed for spatio-temporal activity detection on unconstrained videos~\cite{rizve_gabriella_2021,yu_cmu_2020,liu_argus_2020,chang_mmvg-inf-etrol_2019, yu_training-free_2019, yu_traffic_2018}. 
Generally, theses systems first generates activity proposals and then feeds them to classification models.
Since there have been a variety of video classification networks~\cite{tran_closer_2018, lin_tsm_2019, feichtenhofer_x3d_2020}, the major focus is on the paradigm of proposals and the generation algorithm.
In \cite{liu_argus_2020, chang_mmvg-inf-etrol_2019}, a detection and tracking framework is employed to extract whole object tracklets as tubelets, where temporal localization is required.
In \cite{rizve_gabriella_2021}, an encoder-decoder network is used to generate localization masks on fixed-length clips for tubelet proposal extraction, which has varied spatial locations in different frames.


\section{Method}

\subsection{Activity Detection Task}

In this paper, we tackle the activity detection task in unconstrained videos which are untrimmed and with large field-of-views.
Given an untrimmed video stream $\mathcal{V}$, the system $\mathcal{S}$ should identify a set of activity instances $\mathcal{S}(\mathcal{V})=\{A_i\}$.
Each activity instance is defined by a three-tuple $A_i=(T_i, L_i, C_i)$, referring to an activity of type $C_i$ occurs at temporal window $T_i$ with spatial location $L_i$.
$L_i$ contains the precise location of $A_i$ in each frame, forming a tube in the timeline.
As such, activity detection can often be decomposed into three aspects, i.e., temporal localization ($T_i$), spatial localization ($L_i$), and action classification ($C_i$). 

Each of the three aspects poses unique challenges to the video understanding system. Due to its multi-dimensional nature, it remains hard to define and build a useful activity detection system under the strict setting. Therefore, we also evaluates with some loosened requirements.

\paragraph{Strict Setting}

All activity types are defined as atomic activities with clear temporal boundaries and spatial extents.
The evaluation metric performs bipartite matching between predictions and ground truths.

\paragraph{Loosened Setting}

Activity types are either atomic activities within a temporal window (e.g. standing up) or continuous repetitive activities that can be cut into multiple identifiable windows (e.g. walking).
The evaluation metric allows multiple non-overlapping predictions to be matched with one ground truth.

\subsection{Argus++ System}

The architecture of the proposed \textit{Argus++} system is shown in Figure \ref{fig/arch}.
To tackle the task of activity detection, we adopt an intermediate concept of \textit{spatio-temporal cube proposal} with a much simpler definition than an activity instance:
\begin{equation}
    p_i = (x_0^i, x_1^i, y_0^i, y_1^i, t_0^i, t_1^i)
\end{equation}
This six-tuple design relieves the localization precision and caters modern action classification models which works on fixed-length clips with fixed spatial window.

For an input video stream, the system first generates candidate proposals with frame-wise information such as detected objects, which will be covered in Section \ref{sec/prop_gen}.
These proposals are filtered with a background subtraction model as detailed in Section \ref{sec/prop_fil}.
Then, action recognition models described in Section \ref{sec/act_rec} are applied on the proposals to predict per-class confidence scores.
Finally, Section \ref{sec/act_ded} introduces the post-processing stage to merge and filter the proposals with scores and generate final activity instances.

\subsection{Proposal Generation} \label{sec/prop_gen}

Starting this section, we introduce each of the components of \textit{Argus++}.
The system begins by generating a set of cube proposals.
They are generated based on information from frame-level object detection with multiple object tracking methods. 
Cubes are sampled densely in the timeline with refined spatial locations.

\paragraph{Detection and Tracking}
To conduct activity recognition, we first locate the candidate objects (in most cases, person and vehicle) in the video. 
For each selected frame $F_i$, we apply an object detection model to get objects $O_i=\{o_{i, j} \mid j=1, \cdots, n_i\}$ with object types $c_{i, j}$ and bounding boxes $(x_0, x_1, y_0, y_1)_{i, j}$.
Objects are detected in a stride of every $S_\mathit{det}$ frames.
A multiple object tracking algorithm is applied on the detected objects to assign track ids to each of them as $\mathit{tr}_{i, j}$.

\begin{figure}[!ht]
	\centering
	\includegraphics[width=0.6\linewidth]{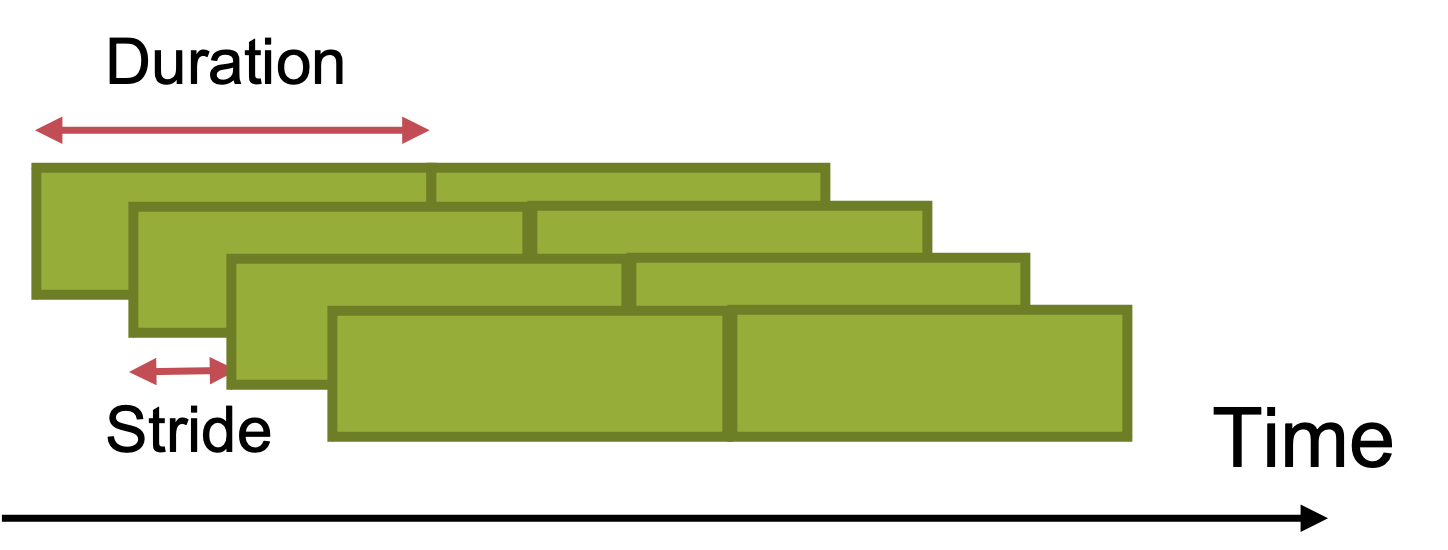}
	\caption{Dense Overlapping Proposals}
	\label{fig/cube}
\end{figure}

\paragraph{Proposal Sampling}
To sample proposals on untrimmed videos without breaking the completeness of any activity instances, we propose a dense overlapping proposals sampling algorithm. 
As illustrated in Figure \ref{fig/cube}, this method ensures coverage of activities occurring at any time, with no hard boundaries.
Two parameters, duration $D_\mathit{prop}$ and stride $S_\mathit{prop}$, controls the sampling process.
Each proposal contains a temporal window of $D_\mathit{prop}$ frames.
New proposals are generated every $S_\mathit{prop} \le D_\mathit{prop}$ frames, possibly with overlaps. 
Generally, non-overlapping proposal system can be treated as a degraded case when $S_\mathit{prop} = D_\mathit{prop}$.

\paragraph{Proposal Refinement}
To generate proposals in a temporal window from $t_0$ to $t_1=t_0 + D_\mathit{prop}$, we select seed track ids $\mathit{Tr}_{t_c}$ from the central frame $t_c = \lfloor\frac{t_0 + t_1}{2}\rfloor$.
Their bounding boxes are enlarged as the union across the temporal window
\begin{equation}
    \begin{aligned}
        (x_0, x_1, y_0, y_1)_k =& \bigcup(\{(x_0, x_1, y_0, y_1)_{i, j} \mid\\
        & t_0 \le i \le t_1, \mathit{tr}_{i, j} = \mathit{tr}_{t_c, k}\})\\
        k =& 1, \cdots, n_{t_c}
    \end{aligned}
\end{equation}
This algorithm is robust through identity switch in the tracking algorithm as it uses the stable seeds from the central frame.
It also ensures the coverage of moving objects by enlarging the bounding box when it's successfully tracked.
This design is helpful for efficiency optimization by allowing a large detection stride $S_\mathit{det}$.
When later applied for activity recognition, the bounding box can be further enlarged for a fixed rate $R_{enl}$ to include spatial context and compensate for missed tracks.

\begin{figure}[!t]
	\centering
	\includegraphics[width=\linewidth]{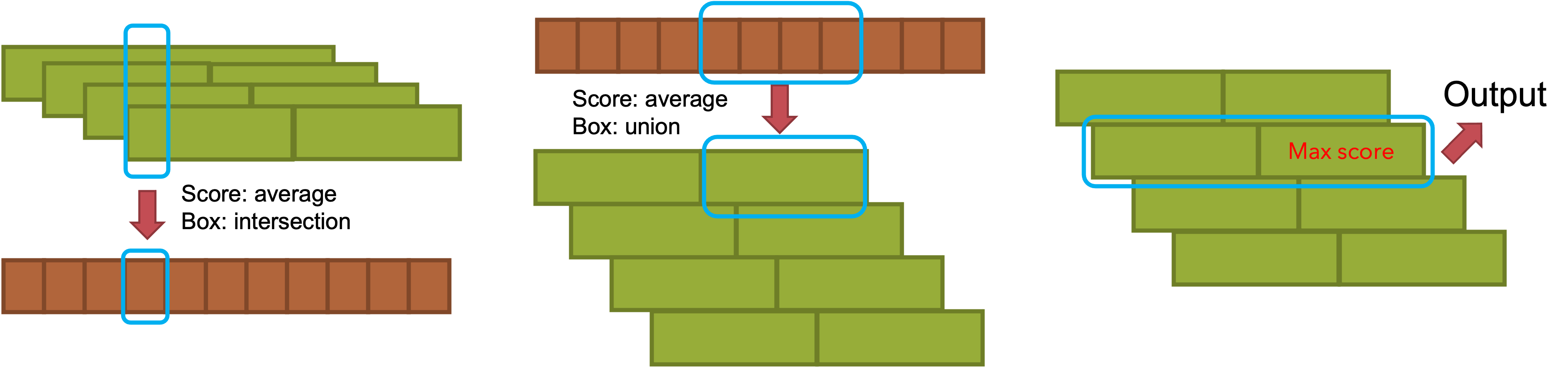}
	\caption{Deduplication Algorithm for Overlapping Proposals}
	\label{fig/merge}
\end{figure}

\subsection{Proposal Filtering} \label{sec/prop_fil}

For now, the proposal generation pipeline applies a frame-wise object detection with slight aid of tracking information. 
The motion information of video is not yet explored.
To produce high quality proposals, we apply a proposal filtering algorithm to eliminate the proposals that are unlikely to contain activities.

\paragraph{Foreground Segmentation}
For each proposal, a foreground segmentation algorithm is implemented to generate a binary mask for every $S_{bg}$ frames for each video clip. 
We average the value of pixel masks in its cube to get its foreground score $f_i$. 
For proposals generated by object type $c$, those proposals with $f_i \le F_c$ will be filtered out.
The threshold $F_c$ is determined by allowing up to $P_\mathit{pos}$ true proposals to be filtered out.

\paragraph{Label Assignment}
To determine the above threshold and to train the activity recognition module, we need to assign labels for each generated proposal according to the ground truth activity instances.
We first convert the annotation of activity instances into the cube format, denoted as ground truth cubes, by performing dense sampling of duration $D_\mathit{prop}$ and stride $S_\mathit{prop}$ within each instance.
For each proposal, we estimate the spatial intersection-over-union (IoU) between it and ground truth cubes in the same temporal window. 
Then we follow Faster R-CNN~\cite{ren_faster_2015} in the assignment process:
\begin{itemize}
    \item For each ground truth cube, assign it to the proposal with the highest score above $S_{low}$.
    \item For each proposal, assign it with each ground truth cube with score above $S_{high}$.
    \item For each proposal, assign it as negative if all scores are below $S_{low}$.
\end{itemize}
$S_{high}$ and $S_{low}$ are the high and low thresholds.
Through this algorithm, each proposal may be assigned one or more positive labels, a negative label, or nothing.
Those assigned nothing are redundant detections which will not be used in classifier training.

\paragraph{Proposal Evaluation}
To measure the quality of proposals before and after the filtering, we need a method for proposal evaluation.
This can be achieved by assuming a perfect classifier in the activity recognition part, so the final metrics reflects the upper bound performance with current proposals.
To do this, we simply use the assigned labels as the classification outputs and pass through the deduplication algorithm covered later.
To further measure other properties of the generated proposals, we can only pass through a subset of them, such as only those with spatial IoU against ground truth above 0.5.

\begin{table*}[htb]
\centering
\caption{CVPR 2021 ActivityNet Challenge\protect\footnotemark ActEV SDL Unknown Facility Evaluation}
\label{exp/activitynet21}
\begin{tabular}{@{}lccc@{}}
\toprule
System/Team&$\mathit{nAUDC}@0.2T_\mathit{fa}\downarrow$&Mean$P_{miss}@0.02T_\mathit{fa}\downarrow$&Relative Processing Time\\ \midrule
    \textbf{Argus++ (Ours)} & \textbf{0.3535} & \textbf{0.5747} & 0.576  \\
    UMD\_JHU & \underline{0.4232} & 0.6250 & 0.345  \\ 
    IBM-Purdue & 0.4238 & 0.6286 & 0.530 \\
    UCF & 0.4487 & \underline{0.5858} & 0.615 \\
    Visym Labs & 0.4906 & 0.6775 & 0.770 \\
    MINDS\_JHU & 0.6343 & 0.7791 & 0.898 \\
    \bottomrule
\end{tabular}
\end{table*}

\footnotetext{\url{http://activity-net.org/challenges/2021/challenge.html}}

\begin{table*}[htb]
\centering
\caption{NIST ActEV'21 SDL\protect\footnotemark Known Facility Evaluation}
\label{exp/sdl_kf}
\begin{tabular}{@{}lccc@{}}
\toprule
System/Team&$\mathit{nAUDC}@0.2T_\mathit{fa}\downarrow$&Mean$P_{miss}@0.02T_\mathit{fa}\downarrow$&Relative Processing Time\\ \midrule
    \textbf{Argus++ (Ours)} & \textbf{0.1635} & \textbf{0.3424} & 0.413  \\
    UCF & \underline{0.2325} & \underline{0.3793} & 0.751  \\ 
    UMD & 0.2628 & 0.4544 & 0.380 \\
    IBM-Purdue & 0.2817 & 0.4942 & 0.631 \\
    Visym Labs & 0.2835 & 0.4620 & 0.721 \\
    UMD-Columbia & 0.3055 & 0.4716 & 0.516 \\
    UMCMU & 0.3236 & 0.5297 & 0.464 \\
    Purdue & 0.3327 & 0.5853 & 0.131 \\
    MINDS\_JHU & 0.4834 & 0.6649 & 0.967 \\
    BUPT-MCPRL & 0.7985 & 0.9281 & 0.123 \\
    \bottomrule
\end{tabular}
\end{table*}

\footnotetext{\url{https://actev.nist.gov/sdl\#tab_leaderboard}}

\subsection{Activity Recognition} \label{sec/act_rec}
In this section, we will elaborately introduce our action recognition modules. Given the input proposal of an activity instance $p_i$, our action recognition model $\mathbb{V}$ will give out the confidence vector $c_i$:
\begin{equation}
    \mathbb{V}(p_i) = c_i = \{c^1_i,c^2_i,...c^n_i\}
\end{equation}
Where n represents the number of target actions, and $c_i \in \mathbb{R}^{n}$. Limited by GPU memory size and temporal length settings of pretrained weights, we need to select t frames out of $t_1^i-t_0^i$ samples from the activity instance. To do this, we strictly followed the sparse-sampling strategy mentioned in \cite{wang_temporal_2016} for both training and inference stage. To be specific, the video is evenly separated into t segments. From each segment, 1 frame will be randomly selected to generate the sampled clip. 

To transform the action recognition modules from previous multi-class task to the realm of multi-label recognition, we modified the loss function for optimization. Instead of traditional cross entropy loss (XE), we implemented a weighted binary cross entropy loss (wBCE). In which, two weight parameters are adopted, the activity-wise weight $\text{W}_a = \{w_a^1,w_a^2,...,w_a^n\}$ and the positive-negative weight $\text{W}_p = \{w_p^1,w_p^2,...,w_p^n\}$. $\text{W}_a$ balances the training samples of different activities and $\text{W}_p$ balances the positive and negative samples of a specific activity. 
With the aligned label sequence of $i^{th}$ instance represented as $Y_i=\{y_i^1,y_i^2,...,y_i^{n}\} \in \mathbb{R}^n$. The calculation of $w_a^{c}$ is derived as:
\begin{equation}
    \hat{w}_a^c =\frac{1}{\sum_{i\in [I]}y_i^c}
\end{equation}
\begin{equation}
    w_a^{c} = n \times \frac{\hat{w}_a^c}{\sum_{c\in[n]}\hat{w}_a^c}
\end{equation}
And the derivation of $w_p^{c}$ is:
\begin{equation}
    w_p^{c} =\frac{\sum_{i\in [I]}\mathbf{1}_{y_i^c=0}}{\sum_{i\in [I]}y_i^c}
\end{equation}
In which, $[I]$ represents all input instances, and $[n]$ represent all target activities.
Compared with vanilla BCE loss, we found wBCE loss can significantly improve the final performance on internal validation set.

Furthermore, we tried multiple action recognition modules and made late fusion action-wisely according to the results on the validation set. We found each classifier does show superiority on certain actions. Through the feedback from the online leaderboard,  such fusion strategy can improve the final performance with noticeable margins.

\begin{table*}[htb]
\centering
\caption{NIST ActEV'21 SDL Unknown Facility Evaluation}
\label{exp/sdl_uf}
\begin{tabular}{@{}lccc@{}}
\toprule
System/Team&$\mathit{nAUDC}@0.2T_\mathit{fa}\downarrow$&Mean$P_{miss}@0.02T_\mathit{fa}\downarrow$&Relative Processing Time\\ \midrule
    \textbf{Argus++ (Ours)} & \textbf{0.3330} & \underline{0.5438} & 0.776  \\
    UCF & \underline{0.3518} & \textbf{0.5372} & 0.684  \\ 
    IBM-Purdue & 0.3533 & 0.5531 & 0.575 \\
    Visym Labs & 0.3762 & 0.5559 & 1.027 \\
    UMD & 0.3898 & 0.5938 & 0.515 \\
    UMD-Columbia & 0.4002 & 0.5975 & 0.520 \\
    UMCMU & 0.4922 & 0.6861 & 0.614 \\
    Purdue & 0.4942 & 0.7294 & 0.239 \\
    MINDS\_JHU & 0.6343 & 0.7791 & 0.898 \\
    \bottomrule
\end{tabular}
\end{table*}



\subsection{Activity Deduplication} \label{sec/act_ded}

\paragraph{Overlapping Instances}

As the system generates overlapping proposals, it could have duplicate predictions for some of the proposals.
This would result in a large amount of false alarms unless we deduplicate them.
Figure \ref{fig/merge} is a diagram for our deduplication algorithm which applies to each activity type with all proposals: 
\begin{enumerate}
    \item Split the overlapping cubes of duration $D_\mathit{prop}$ and stride $S_\mathit{prop}$ into non-overlapping cubes of duration $S_\mathit{prop}$.
    An output cube relies on all original cubes in the temporal window, with an averaged score and an intersected bounding box.
    \item Merge the non-overlapping cubes of duration $S_\mathit{prop}$ back into $\lfloor\frac{D_\mathit{prop}}{S_\mathit{prop}}\rfloor$ groups of non-overlapping cubes of duration $D_\mathit{prop}$.
    An output cube is merged from $\lfloor\frac{D_\mathit{prop}}{S_\mathit{prop}}\rfloor$ cubes with an averaged score and the union of bounding boxes.
    \item Select the group where the maximum score resides.
\end{enumerate}

The deduplication algorithm performs an interpolation upon the overlapping cubes.
Each group in step 3 contains information from every classification results, maximizing the information utilization.

\paragraph{Adjacent Instances}
The above deduplication process only transforms overlapping instances to non-overlapping instances with the same duration.
This would be sufficient under the \textit{Loosened Setting}, where multiple predictions are allowed for each activity.
No threshold would be needed to truncate low-confidence predictions as this happens automatically during the ground-truth matching process.

However, for the \textit{Strict Setting}, we need to further merge adjacent cubes into integrate instances.
Currently we adopt a simple yet effective algorithm, by simply merging adjacent cubes where all of them have confidence score above $S_{merg}$. 
The merged instance needs to be longer than $L_{merg}$ to be kept in the final output.

\section{Experiments}

\begin{table*}[htb]
\centering
\caption{NIST TRECVID 2021 ActEV Evaluation~\cite{2021trecvidawad, yu_cmu_2021}}
\label{exp/trecvid21}
\begin{tabular}{@{}lccc@{}}
\toprule
System/Team&$\mathit{nAUDC}@0.2T_\mathit{fa}\downarrow$&Mean  $P_{miss}@0.15T_\mathit{fa}\downarrow$&Mean $wP_{miss}@0.15R_\mathit{fa}\downarrow$ \\\midrule
    \textbf{Argus++ (Ours)} & \textbf{0.39607} & \textbf{0.30622} & \underline{0.81080}  \\
    BUPT & \underline{0.40853} & \underline{0.32489} & \textbf{0.79798}  \\ 
    UCF & 0.43059 & 0.34080 & 0.86431 \\
    M4D & 0.84658 & 0.79410 & 0.88521 \\
    TokyoTech\_AIST & 0.85159 & 0.81970 & 0.94897 \\
    Team UEC & 0.96405 & 0.95035 & 0.95670 \\
    \bottomrule
\end{tabular}
\end{table*}

\begin{table*}[htb]
\centering
\caption{NIST TRECVID 2020 ActEV Evaluation~\cite{awad_trecvid_2021, yu_cmu_2020}}
\label{exp/trecvid20}
\begin{tabular}{@{}lccc@{}}
\toprule
System/Team                                     & $\mathit{nAUDC}@0.2T_\mathit{fa}\downarrow$ & Mean $P_{miss}@0.15T_\mathit{fa}\downarrow$ & Mean $wP_{miss}@0.15R_\mathit{fa}\downarrow$ \\ \midrule
\textbf{Argus++ (Ours)} & \textbf{0.42307} & \textbf{0.33241} & \textbf{0.80965} \\
UCF                                      & \underline{0.54830}                           & 0.50285                           & \underline{0.83621}                           \\
BUPT-MCPRL                               & 0.55515                           & \underline{0.48779}                           & 0.84519                           \\
TokyoTech\_AIST                          & 0.79753                           & 0.75502                           & 0.87889                           \\
CERTH-ITI                                & 0.86576                           & 0.84454                           & 0.88237                           \\
Team UEC                                 & 0.95168                           & 0.95329                           & 0.98300                           \\
Kindai\_Kobe                             & 0.96267                           & 0.95204                           & 0.93905                           \\ \bottomrule
\end{tabular}
\end{table*}

\subsection{Implementation Details}

In \textit{Argus++}, we apply Mask R-CNN~\cite{he_mask_2020} with a ResNet-101~\cite{he_deep_2016} backbone from Detectron2~\cite{wu2019detectron2} pre-trained on the Microsoft COCO dataset~\cite{lin2014microsoft} as the object detector, with $S_\mathit{det}=8$. Only person, vehicle, and traffic light classes are selected.
For the tracking algorithm, we apply the work in \cite{vedaldi_towards_2020} and reuse the region-of-interest from the ResNet backbone as in \cite{yu_zero-virus_2020, qian_electricity_2020}.

The proposals are generated with $D_\mathit{prop}=64$ and $S_\mathit{prop}=16$.
The labels are assigned with $S_{high}=0.5$ and $S_{low}=0$.
The proposal filter is set with a tolerance of $P_\mathit{pos}=0.05$.

For activity classifiers, we adopted multiple state-of-the-art models including R(2+1)D~\cite{tran_closer_2018}, X3D~\cite{feichtenhofer_x3d_2020}, and Temporal Relocation Module (TRM)~\cite{qian2022trm}.
During training procedure, frames are cropped with jittering~\cite{wang_temporal_2016} and enlarged with $R_{enl}=0.13$. For X3D and TRM, we trained modules with weights pre-trained on Kinetics~\cite{kay_kinetics_2017}. For R(2+1)D modules, we trained modules with weighst pre-trained on IG65M~\cite{ghadiyaram_large-scale_2019}. 
We fused confidence scores from these models according to their performance on the validation set.

\subsection{Evaluation Protocols}

To measure the performance, efficiency, and generalizability of \textit{Argus++}, we evaluate it across a series of public benchmarks.
\textit{Argus++} is applied to NIST Activities in Extended Videos (ActEV) evaluations on MEVA~\cite{corona_meva_2021} Unknown Facility , MEVA Known Facility, and VIRAT~\cite{oh_large-scale_2011} settings for surveillance activity detection. 
With slight modifications, it is also tested in the ICCV 2021 ROAD challenge for the action detection task in autonomous driving.

In the NIST evaluations, the metrics~\cite{awad_trecvid_2021} are designed in the \textit{Loosened Setting}, where short-duration outputs are allowed and spatial alignment is ignored.
The idea was that, after processed by the system, there will still be human reviewers to inspect the activity instances with the highest confidence scores for further usages.
The performance is thus measured by the probability of miss detection ($P_{miss}$) of activity instances within a time limit of all positive frames plus $T_\mathit{fa}$ of negative frames, where $T_\mathit{fa}$ is referred to as time-based false alarm rate.
The major metric, $\mathit{nAUDC}@0.2T_\mathit{fa}$, is an integration of $P_{miss}$ on $T_\mathit{fa} \in [0, 0.2]$.

In the ROAD challenge, the \textit{Strict Setting} is adopted by using the mean average precision (mAP) at 3D intersection-over-union (IoU) evaluation metric.
This metric does exact bipartite matching between predictions and ground truth instances, with challenging localization precision requirements.

For metrics in the following tables, $\downarrow$ means lower is better and $\uparrow$ means higher is better. For each metric, the best value is bolded and the second best is underscored. For ongoing public evaluations, the result snapshot at 11/01/2021 is presented.

\subsection{ActEV Sequestered-Data Evaluation}

ActEV Sequestered Data Leaderboards (SDL) are platforms where a system is submitted to run on NIST's evaluation servers.
This submission format prevents access to the test data and measures the processing time with unified hardware platform\footnote{\url{https://actev.nist.gov/pub/Phase3_ActEV_2021_SDL_EvaluationPlan_20210803.pdf}}.
For these evaluations, \textit{Argus++} was trained on MEVA, a large-scale surveillance video dataset with activity annotations of 37 types.
We used 1946 videos in its training release drop 11 as the training set and 257 videos in its KF1 release as validation set.
The optimization target is reaching better performance within 1x real-time.

Table \ref{exp/activitynet21} shows the published results from CVPR 2021 ActivityNet Challenge ActEV SDL Unknown Facility evaluation, where \textit{Argus++} demonstrated around 20\% advantage in $\mathit{nAUDC}@0.2T_\mathit{fa}$ over runner-up system.
The test set of unknown facility is captured with a different setting from MEVA, which challenges the generalization of action detection models.
Table \ref{exp/sdl_kf} shows the ongoing NIST ActEV'21 SDL Known Facility leaderboard, where \textit{Argus++} shows over 40\% advantage in $\mathit{nAUDC}@0.2T_\mathit{fa}$.
The test set of known facility shares a similar distribution with MEVA, where our system learns well and is getting nearer for real-world usages.
Table \ref{exp/sdl_uf} shows the ongoing NIST ActEV'21 SDL Unknown Facility leaderboard continued from ActivityNet, where \textit{Argus++} still holds the leading position with over 5\% advantage in $\mathit{nAUDC}@0.2T_\mathit{fa}$.

\subsection{ActEV Self-Reported Evaluation}

ActEV self-reported evaluations are where only results are submitted and test data is accessible.
This currently includes the annual TRECVID ActEV evaluations on VIRAT.
For TRECVID, we use the official splits of VIRAT for training and validation.

Table \ref{exp/trecvid21} and \ref{exp/trecvid20} shows the leaderboard of 2020 and 2021 NIST TRECVID ActEV Challenge. In 2020, our systems is 22.8\% better in $\mathit{nAUDC}@0.2T_\mathit{fa}$, 33.8\% better in Mean $P_{miss}@0.15T_\mathit{fa}$, and 3.5\% better in Mean-$wP_{miss}@0.15R_\mathit{fa}$ than the runner-up. Although the other competitors improved significantly in 2021, our system still holds the first place with noticeable margins.

\subsection{ROAD Challenge}
Different from previous surveillance action detection benchmarks, the videos of ROAD Challenge\cite{RobotCarDatasetIJRR} are gathered from the point of view of autonomous vehicles. 
It contains 122K frames from 22 annotated videos, where each video is 8 minutes long on average. 
Totally 7K tubes of individual agents are included and each tube consists on average of approximately 80 bounding boxes linked over time. 

Table \ref{exp/road} shows the performance of our system with other competitors. Our system ranks the first with 20\% average mAP.
Although the performance is still far from satisfying in this \textit{Strict Setting}, it demonstrates the capability of \textit{Argus++} in adapting to precise 3D localization and moving camera view points.

\begin{table*}[ht]
\centering
\caption{ICCV 2021 ROAD Challenge Action Detection\protect\footnotemark}
\label{exp/road}
\begin{tabular}{@{}lcccc@{}}
\toprule
System/Team    & Action@0.1 $\uparrow$ & Action@0.2 $\uparrow$ & Action@0.5 $\uparrow$ & \textbf{Average} $\uparrow$ \\ \midrule
\textbf{Argus++ (Ours)} & \textbf{28.54}      & \textbf{25.63}      & 6.98       & \textbf{20.38}   \\
THE IFY        & \underline{28.15}      & \underline{20.97}      & 6.58       & \underline{18.57}   \\
YAAAHO         & 26.81      & 20.40      & \underline{7.02}       & 18.07   \\
hyj            & 26.52      & 20.32      & \textbf{7.05}       & 17.97   \\
3D RetinaNet~\cite{singh_road_2021}   & 25.70      & 19.40      & 6.47       & 17.19   \\
LeeC           & 13.64      & 9.89       & 2.23       & 8.59   \\ \bottomrule
\end{tabular}
\end{table*}

\begin{table*}[ht]
\centering
\caption{Proposal Quality Metrics on VIRAT Validation Set}
\label{tab/prop_qual}
\begin{tabular}{@{}lcccccc@{}}
\toprule
$\mathit{nAUDC}@0.2T_\mathit{fa}$     & \multicolumn{3}{c}{IoU}       & \multicolumn{3}{c}{Reference Coverage}    \\
Threshold & Average & $\ge$ 0 & $\ge$ 0.5 & Average & $\ge$ 0.5 & $\ge$ 0.9 \\  \midrule
Unfiltered Proposals & 0.2358  & 0.0772  & 0.1518    & 0.1562  & 0.1125    & 0.4211    \\
Filtered Proposals   & 0.2352  & 0.0772  & 0.1469    & 0.1563  & 0.1099    & 0.4280    \\ \bottomrule
\end{tabular}
\end{table*}

\footnotetext{\url{https://eval.ai/web/challenges/challenge-page/1059/leaderboard/2748}}

\subsection{Ablation Study}

\paragraph{Coverage of Proposal Formats}

We analyze the coverage of dense spatio-temporal proposals and determines the best hyper-parameters for the proposal format. 
By directly use ground truth cubes as proposals, we estimate the upper bound performance of both overlapping and non-overlapping proposal formats on VIRAT validation set.
The results are shown in Table \ref{tab/overlap}, where non-overlapping proposals shows at least 6.7\% systematic errors while overlapping proposals with duration 64 and stride 16 only has 1.3\%.

\begin{table}[!h]
\centering
\caption{Lower Bounds of $\mathit{nAUDC}@0.2T_\mathit{fa}$ on VIRAT Validation Set with different proposal formats. Italic values are non-overlapping proposals while the others are overlapping proposals. Duration and stride are in the unit of frames.}
\label{tab/overlap}
\begin{tabular}{@{}ccccc@{}}
\toprule
Duration / Stride & 16     & 32   & 64   & 96     \\ \midrule
32                          & 0.0705 & \textit{0.1208} & - & -       \\
64                          & \textbf{0.0127} & 0.0621 & \textit{0.0673} & -  \\
96                          & 0.0275 & 0.0504 & - & \textit{0.0688} \\ \bottomrule
\end{tabular}
\end{table}

\paragraph{Performance of Proposal Filtering}

We examine the quality of the proposals with and without the filter, as shown in Table \ref{tab/prop_sta} and \ref{tab/prop_qual}.
With the proposal evaluation procedure introduced in Section \ref{sec/prop_fil}, the proposals are further filtered by IoU with reference and coverage of reference at levels from 0, 0.1, to 0.9 to calculate partial results.

\begin{table}[!h]
\centering
\caption{Statistics of Proposals on VIRAT Validation Set}
\label{tab/prop_sta}
\begin{tabular}{@{}lcc@{}}
\toprule
Name                    & Unfiltered  & Filtered   \\ \midrule
Number of Proposals         & 211271      & 62831  \\
Positive rate            & 0.1704  & \textbf{0.5204} \\
Rate of unique label & 0.4558 & 0.4415 \\
Rate of two labels & 0.4127 & 0.4252 \\
Rate of three labels & 0.1017 & 0.1060 \\ \bottomrule
\end{tabular}
\end{table}

With the dense cube proposals, the best $\mathit{nAUDC}@0.2T_\mathit{fa}$ we can achieve with a ideal classifier is 0.08, as indicated in the IoU $\ge 0$ column.
The IoU and reference coverage bounded scores are used to measure the spatial matching quality of proposals, as the $\mathit{nAUDC}@0.2T_\mathit{fa}$ does not consider spatial alignments.
We can see that even with a condition of IoU $\ge 0.5$, our proposal can achieve up to 0.15, which indicates the spatial preciseness.
The proposal filter is also proved effective, which removed 70\% of original proposals without dropping the recall level.

The effect of the proposal filter is also evaluate on the SDL, as shown in Table \ref{exp/sdl_prop}.
It not only reduces processing time from 0.925 to 0.582, but also improves $\mathit{nAUDC}@0.2T_\mathit{fa}$ due to reduced false alarms.

\begin{table}[htb]
\centering
\caption{Proposal Filter on NIST ActEV'21 SDL Unknown Facility Micro Set}
\label{exp/sdl_prop}
\begin{tabular}{@{}lcc@{}}
\toprule
Proposal Filter&$\mathit{nAUDC}@0.2T_\mathit{fa}\downarrow$& Processing Time\\ \midrule
    \textbf{Enabled} & \textbf{0.4822} & 0.582  \\ 
    Disabled & 0.5176 & 0.925 \\
    \bottomrule
\end{tabular}
\end{table}


\section{Conclusion}

In this work, we proposed \textit{Argus++}, a robust real-time activity detection system for analyzing unconstrained video streams. We introduced \textit{overlapping spatio-temporal cubes} as an intermediate concept of activity proposals to ensure coverage and completeness of activity detection through over-sampling. The proposed system is able to process unconstrained videos with robust performance across multiple scenarios and real-time effiency on consumer-level hardware. 
Extensive experiments on different surveillance and driving scenarios demonstrated its superior performance in a series of activity detection benchmarks, including CVPR ActivityNet ActEV 2021, NIST ActEV SDL UF/KF, TRECVID ActEV 2020/2021, and ICCV ROAD 2021.

Future works are suggested to focus on extending the current system to more applications, such as action detection in UAV captured videos, first-person human activity understanding, etc. The proposed system could also be extended to end-to-end frameworks for better performance.

\ifwacvfinal
\section{Acknowledgements}
This research is supported in part by the Intelligence Advanced Research Projects Activity (IARPA) via Department of Interior/Interior Business Center (DOI/IBC) contract number D17PC00340. This research is supported in part through the financial assistance award 60NANB17D156 from U.S. Department of Commerce, National Institute of Standards and Technology. This project is funded in part by Carnegie Mellon University’s Mobility21 National University Transportation Center, which is sponsored by the US Department of Transportation.
\fi

\newpage
{\small
\bibliographystyle{ieee_fullname}
\bibliography{ref_zotero.bib, ref_local.bib,ref_related_work.bib}
}

\end{document}